\documentclass[lettersize,journal]{IEEEtran}
\usepackage{amsmath,amsfonts}
\usepackage{algorithmic}
\usepackage{algorithm}
\usepackage{array}
\usepackage[caption=false,font=normalsize,labelfont=sf,textfont=sf]{subfig}
\usepackage{textcomp}
\usepackage{stfloats}
\usepackage{url}
\usepackage{verbatim}
\usepackage[pdftex]{graphicx}
\usepackage[pdftex]{color}
\usepackage{cite}
\usepackage{amssymb,amsopn,bm}
\usepackage{afterpage}
\usepackage{color}
\usepackage{url}
\usepackage{color}
\usepackage[table]{xcolor}
\usepackage{booktabs, multirow}

\begin{document}

\title{Personalized Image Enhancement \\Featuring Masked Style Modeling}

\author{{Satoshi Kosugi and Toshihiko Yamasaki,\IEEEmembership{~Member,~IEEE}}
\thanks{Satoshi Kosugi was with the Department of Information and Communication
Engineering, The University of Tokyo, Bunkyo-ku, Tokyo 113-8656, Japan.
He is now with the Institute of Innovative Research, Tokyo Institute of Technology, Yokohama, Kanagawa 226-8503, Japan
(e-mail: kosugi.s.aa@m.titech.ac.jp).

Toshihiko Yamasaki is with the Department of Information and Communication
Engineering, The University of Tokyo,  Bunkyo-ku, Tokyo 113-8656, Japan.

This work was partially financially supported by JST AIP Acceleration Research JPMJCR22U4 and JSPS KAKENHI Grant Number 22H03640, Japan.

Copyright © 2023 IEEE. Personal use of this material is permitted. However, permission to use this material for any other purposes must be obtained from the IEEE by sending an email to pubs-permissions@ieee.org.}}

\markboth{IEEE TRANSACTIONS ON CIRCUITS AND SYSTEMS FOR VIDEO TECHNOLOGY, VOL. XX, NO. XX, 2023}%
{Satoshi Kosugi and Toshihiko Yamasaki: Personalized Image Enhancement Featuring Masked Style Modeling}


\maketitle

\begin{abstract}
		We address personalized image enhancement in this study,
    where we enhance input images for each user based on the user's preferred images.
    Previous methods apply the same preferred style to all input images ({\it i.e.}, only one style for each user); in contrast to these methods, we aim to achieve content-aware personalization by applying different styles to each image considering the contents.
    For content-aware personalization, we make two contributions.
		First, we propose a method named masked style modeling, which can predict a style for an input image considering the contents by using the framework of masked language modeling.
		Second, to allow this model to consider the contents of images,
		we propose a novel training scheme
		where we download images from Flickr and create pseudo input and retouched image pairs using a degrading model.
    We conduct quantitative evaluations and a user study, and our method trained using our training scheme successfully achieves content-aware personalization; moreover, our method outperforms other previous methods in this field.
		Our source code is available at https://github.com/satoshi-kosugi/masked-style-modeling.

\end{abstract}


\section{Introduction}

\IEEEPARstart{A}{} variety of factors can affect the quality of photographs. To automatically improve the quality of photos, many researchers have been working on image enhancement. However, different users have different preferences for enhanced results; therefore, in this study, we address personalized image enhancement (PIE).
We use a large set of known users' preferred images when training an enhancement model. When testing, the enhancement model transfers an unseen input image to an enhanced image that is personalized for a new user based on the new user's preferred images. Because it is difficult for each new user to provide numerous preferred images, we achieve PIE using only a limited number ({\it i.e.}, a few dozen) of preferred images.

\begin{figure}[t]
\centering
\includegraphics[width=1\hsize]{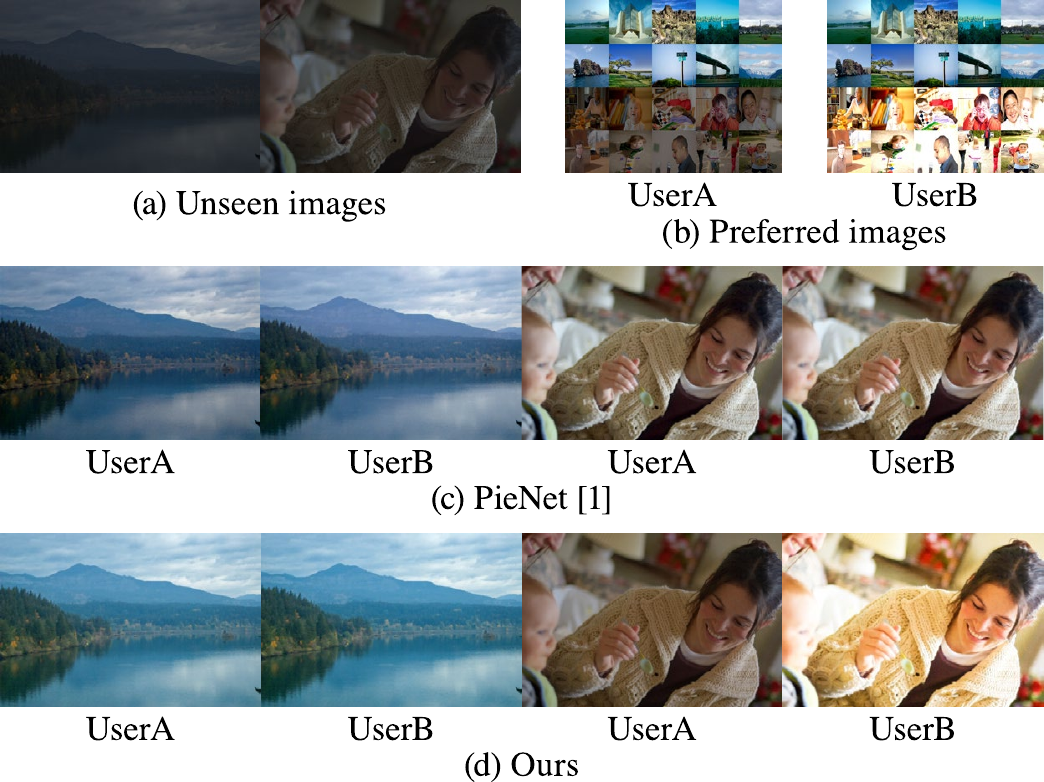}
\caption{Visualization of content-aware PIE.
We consider the case where two users have the same preferences
for landscapes and different preferences for portraits as shown in (b).
When a previous PIE method~\cite{kim2020pienet} enhances unseen images in (a),
as with personalization results for the landscape, results for the portrait have only a slight difference for each user as shown in (c).
Compared to this method, our method can generate almost the same results for the landscape
and different results for the portrait as shown in (d).
}
\label{shin_fig1}
\end{figure}

\begin{figure*}[t]
\centering
\includegraphics[width=1\hsize]{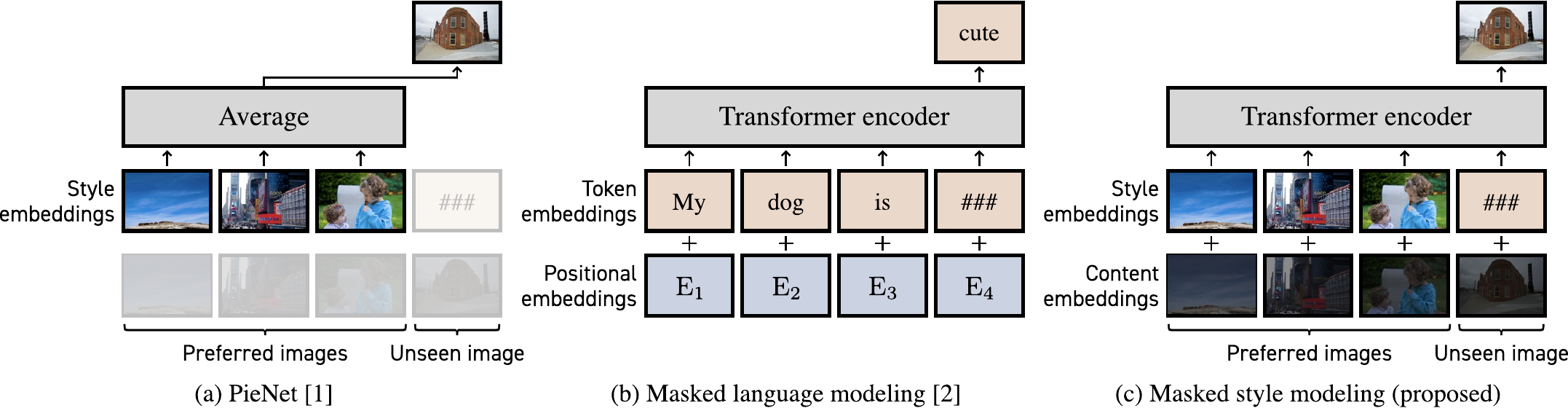}
  \caption{Comparison with a previous method, PieNet~\cite{kim2020pienet}. In PieNet, the average of the style embeddings is used as the user's preference vector, and the same preference vector is applied to all unseen images without considering the contents. For content-aware PIE, we propose masked style modeling inspired by masked language modeling~\cite{kenton2019bert}, where the Transformer encoder receives style and content embeddings instead of token and positional embeddings, and a style embedding of an unseen image is predicted considering the contents.}
\label{figure1}
\end{figure*}

A simple approach for PIE is to pre-train general image enhancement models using all known users' preferred images and subsequently fine-tune them with a new user's preferred images.
However, these models are not suited to be fine-tuned with a limited number of preferred images because of a large number of model parameters,
and they require fine-tuning for each user, which severely limits the scalability to a large number of users.
To solve these problems, some models specialized for PIE have been proposed~\cite{kim2020pienet,bianco2020personalized}.
These models can personalize the enhanced results based on preference vectors.
For a new user, they only need to estimate a preference vector from the preferred images,
and fine-tuning of the entire models is not required.

To generate results that are more preferred by users,
it is important to achieve \textbf{content-aware PIE}.
Because user preferences should depend on the contents of images,
we should apply different styles to each image considering the contents.
However, the previous PIE methods cannot consider the contents
for the following two reasons.
First, the previous PIE models are designed to apply the same style to all images ({\it i.e.}, only one style for each user).
Personalized images have the same style regardless of the contents.
Second, the previous PIE models are trained using images enhanced by rule-based enhancement methods.
Because the rule-based methods apply the same effect to all images and do not consider the contents,
a PIE model trained with these images cannot achieve content-aware PIE.
To show that a previous PIE method cannot achieve content-aware PIE,
we consider the case where two users have the same preferences
for landscapes and different preferences for portraits as shown in Figure~\ref{shin_fig1}(b).
When a previous PIE method enhances unseen images in Figure~\ref{shin_fig1}(a),
as with personalization results for a landscape, results for a portrait have only a slight difference for each user (Figure~\ref{shin_fig1}(c)).
As shown in this result, the previous PIE method cannot achieve content-aware PIE.

To solve these problems and achieve content-aware PIE,
we make two contributions in this study.
First, we propose a model called masked style modeling, a model inspired by masked language modeling in BERT~\cite{kenton2019bert}.
In masked language modeling, the Transformer encoder receives token embeddings and positional embeddings of each word, after which the masked token is predicted considering the context.
Masked language modeling has shown that the Transformer encoder can consider complex potential relations in input sets, and we found this property to be suitable for content-aware PIE. In masked style modeling, token and positional embeddings are replaced by style and content embeddings of images, and the Transformer encoder predicts a style embedding of an unseen image.
Figure~\ref{figure1} shows the comparison with a previous PIE method.
Note that our model receives a new user's preferred images as inputs and
does not require any fine-tuning processes.

Second, we propose a novel training scheme to train masked style modeling.
While a previous PIE method~\cite{kim2020pienet} uses images enhanced by rule-based enhancement methods,
we train masked style modeling using real users' preferred images.
To this end, we use images uploaded by users on Flickr,
which are retouched by the users based on their contents and suitable for content-aware PIE.
Because original and retouched image pairs are needed to train enhancement models,
we train a degrading model and create pseudo original and retouched image pairs.
We show the personalization results by our masked style modeling trained with our training scheme in Figure~\ref{shin_fig1}(d).
The Transformer encoder can capture the relationship between the contents,
and our model is trained using multiple real users' preferred images;
therefore, content-aware PIE is achieved.

In our experiments, we compare the proposed method with existing methods through quantitative evaluations and a user study.
Our method outperforms existing PIE models and fine-tuned general image enhancement models in both quantitative evaluations and the user study. Visualized attentions of the Transformer encoder show that our method can consider the contents of the images and that our training scheme is essential for content-aware PIE.

The following are the main contributions of this study:
\begin{itemize}
\item To achieve content-aware PIE, we propose masked style modeling that can predict the preferred style considering the contents of images.
\item We propose a novel training scheme where we train our model using multiple users' preferred images on Flickr.
\item Our method outperforms the previous methods by considering the contents.
\end{itemize}

\section{Related Works}
\subsection{General Image Enhancement}
General image enhancement models learn translation using a large number of original and retouched images.
Most of the recent methods
are based on convolutional neural networks (CNNs);
Yan et al.~\cite{yan2016automatic} were the first to apply CNNs to image enhancement. Wang et al.~\cite{wang2019underexposed} developed a novel loss function that enables a spatially smooth enhancement.
Moran et al.~\cite{moran2020deeplpf} designed local parametric filters for a lightweight model.
Kim et al.~\cite{kim2020global} combined global and local enhancement models.
He et al.~\cite{he2020conditional} reproduced image-processing operations using a multilayer perceptron.
Wang et al.~\cite{wang2022neural} improved He et al.'s method for sequential image retouching.
Kosugi et al.~\cite{kosugi2020unpaired} utilized image editing software for artifact-free enhancement.
Dhara et al.~\cite{dhara2021exposedness} proposed a structure-aware exposedness estimation procedure for noise-suppressing enhancement.
Zhao et al.~\cite{zhao2021retinexdip} applied deep image prior to image enhancement.
Li et al.~\cite{li2021low} used a recursive unit to repeatedly unfold the input image for feature extraction.
Afifi et al.~\cite{afifi2021learning} proposed a coarse-to-fine framework to enhance over- and under-exposed images.
Kim et al.~\cite{kim2021representative} developed representative color transform for details and high capacity.
Zhao et al.~\cite{zhao2021deep} adopted invertible neural networks for bidirectional feature learning.
Xu et al.~\cite{xu2022structure} presented a structure-texture aware network to fully consider the global structure and local detailed texture.
Liang et al.~\cite{liang2022self} proposed a self-supervised learning framework.
For real-time image enhancement, Gharbi et al.~\cite{gharbi2017deep} applied a bilateral grid~\cite{chen2007real};
Zeng et al.~\cite{zeng2022learning}, Wang et al.~\cite{wang2021real}, Yang et al.~\cite{yang2022adaint}, and Yang et al.~\cite{yang2022seplut} used
3D lookup tables;
while Zhang et al.~\cite{zhang2021star} used a Transformer~\cite{vaswani2017attention}.

These models can be personalized for a new user by pre-training them with known users' preferred images and fine-tuning them with the new user's preferred images, but
they are not suited to be fine-tuned with a limited number of preferred images because of a large number of model
parameters, and they require fine-tuning for each user, which
severely limits the scalability to a large number of users.

\subsection{Personalized Image Enhancement}
The goal of PIE is to enhance unseen images based on the user's preference. Kang et al.'s method~\cite{kang2010personalization} asked a new user to retouch representative images and applied the retouching parameters of the most similar image to an unseen image. Caicedo et al.~\cite{caicedo2011collaborative} applied collaborative filtering to utilize other users' preferences. These methods can enhance unseen images based on the user's preference but can only use traditional enhancement techniques such as gamma-correction and S-curve.

Deep learning--based methods have been proposed to handle more complex mappings. Kim et al.~\cite{kim2020pienet} trained an embedding network and preference vectors;
subsequently, an encoder-decoder model was trained to generate personalized results based on the preference vectors. Bianco et al.~\cite{bianco2020personalized} trained a network that outputs control points for splines and injected a user's profile into the network for personalization. Song et al.~\cite{song2021starenhancer} proposed StarEnhancer for style-aware enhancement, which can be applied to PIE by considering the preferred images as the target style.
For a new user, these methods only need to estimate a preference vector from the preferred images, and fine-tuning of the entire models is not required.
However, the same preference vector is applied to all images; therefore, these methods cannot achieve content-aware PIE.

\begin{figure*}[t]
\centering
\includegraphics[width=1\hsize]{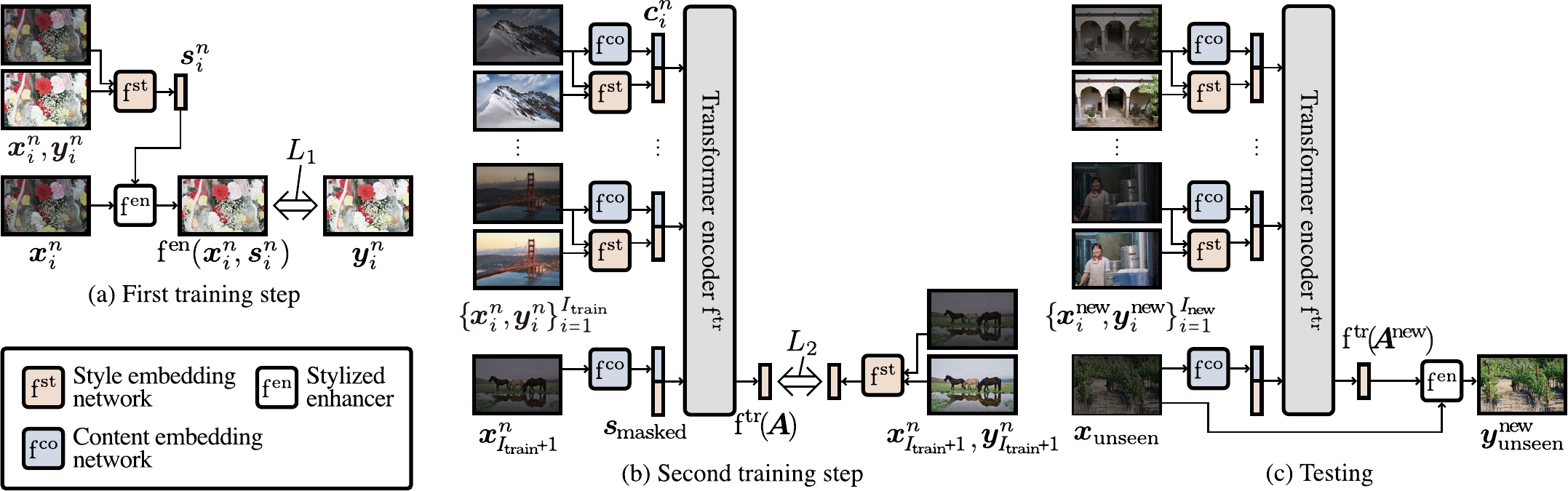}
\caption{Overview of our method.
Our method consists of four networks: a style embedding network ${\rm f^{st}}$,
a content embedding network ${\rm f^{co}}$,
a Transformer encoder ${\rm f^{tr}}$, and a stylized enhancer ${\rm f^{en}}$.
Our training process is divided into two steps.
In the first training step (a), the style embedding network ${\rm f^{st}}$ and the stylized enhancer ${\rm f^{en}}$ are trained.
${\rm f^{st}}$ extracts a style embedding ${\bm s}^n_i = {\rm f^{st}}({\bm y}^n_i) - {\rm f^{st}}({\bm x}^n_i)$,
and ${\rm f^{en}}$ enhances ${\bm x}^n_i$ based on ${\bm s}^n_i$ to predict ${\bm y}^n_i$.
In the second training step (b), the content embedding network ${\rm f^{co}}$ and the Transformer encoder ${\rm f^{tr}}$ are trained.
${\rm f^{co}}$ extracts a content embedding ${\bm c}^n_i = {\rm f^{co}}({\bm x}^n_i)$,
and ${\rm f^{tr}}$ predicts a style embedding for the $(I_{\rm train}+1)$th image
based on the content and style embeddings of $I_{\rm train}$ image pairs.
When testing, a style embedding for an unseen input image ${\bm x}_{\rm unseen}$ is predicted
by ${\rm f^{tr}}$,
and the enhanced result is obtained using ${\rm f^{en}}$.
}
\label{proposed}
\end{figure*}

\section{Masked Style Modeling}
Our goal is to achieve content-aware PIE. When training our enhancement model, we use $N$ known users' preferred image sets $\{{\mathcal S}^n\}_{n=1}^N$, where ${\mathcal S}^n=\{{\bm x}^n_i, {\bm y}^n_i\}_{i=1}^{I_n}$ is the $n$th user's preferred image set, ${\bm x}^n_i$ is a original image, and ${\bm y}^n_i$ is a retouched image. When testing, our enhancement model receives a new user's preferred image set ${\mathcal S}^{\rm new}=\{{\bm x}^{\rm new}_i, {\bm y}^{\rm new}_i\}_{i=1}^{I_{\rm new}}$ and transfers an unseen input image ${\bm x}_{\rm unseen}$ to an enhanced result ${\bm y}^{\rm new}_{\rm unseen}$, which is personalized for the new user.
$I_{\rm new}$ can be different for each new user, but
because it is difficult for each new user to provide numerous preferred images, we limit $I_{\rm new}$ to a few dozen.

\vspace{2mm}
\noindent
{\bf PieNet}~
To describe our motivation, we first introduce a baseline method, PieNet~\cite{kim2020pienet}.
In PieNet, the authors designed a style embedding network ${\rm f^{st}}$ and a stylized enhancer ${\rm f^{en}}$. ${\rm f^{st}}$ is a ResNet-18~\cite{he2016deep} followed by a global average pooling layer, and ${\rm f^{en}}$ is a modified U-net~\cite{ronneberger2015u} where preference vectors can be inserted into the skip connections. They first define preference vectors ${\bm v}^n$ for each user; ${\rm f^{st}}$ and ${\bm v}^n$ are simultaneously optimized by minimizing the following loss,
\begin{equation}
\label{pienetloss1} L^{\rm pienet}_1 = \Bigl[\bigl|\bigl|{\rm f^{st}}({\bm y}^n_i) - {\bm v}^n||^2_2 - ||{\rm f^{st}}({\bm y}^{n'}_{i'}) - {\bm v}^n\bigr|\bigr|^2_2 + \alpha\Bigr]_+,
\end{equation}
where $n\neq n'$, $[]_+$ is a rectifier, and $\alpha$ is a margin. By minimizing this loss, ${\rm f^{st}}({\bm y}^n_i)$ and ${\bm v}^n$ get close, and ${\rm f^{st}}({\bm y}^{n'}_{i'})$ and ${\bm v}^n$ become distant; as a result, the preferred images and the preference vectors are clustered by users. Subsequently, the authors trained ${\rm f^{en}}$ so that the enhanced results are personalized based on ${\bm v}^n$,
\begin{equation}
L^{\rm pienet}_2 = {\rm loss^{pienet}}({\bm y}^n_i, {\rm f^{en}}({\bm x}^n_i, {\bm v}^n)),
\end{equation}
where ${\rm loss^{pienet}}$ is a weighted sum of color, perceptual, and total variation losses. When testing, the preference vector for a new user ${\bm v}^{\rm new}$ is estimated by averaging the embeddings of the preferred images,
\begin{equation}
{\bm v}^{\rm new} = 1/I_{\rm new} \sum\nolimits_{i=1}^{I_{\rm new}} {\rm f^{st}}({\bm y}^{\rm new}_i),
\label{average}
\end{equation}
and an unseen image ${\bm x}_{\rm unseen}$ is enhanced as
\begin{equation}
{\bm y}^{\rm new}_{\rm unseen} = {\rm f^{en}}({\bm x}_{\rm unseen}, {\bm v}^{\rm new}).
\end{equation}
This method can personalize the output based on the preferred images ${\bm y}^{\rm new}_i$
without any fine-tuning processes
but does not consider the contents because the same ${\bm v}^{\rm new}$ is applied to all unseen images.

\begin{figure}[t]
\centering
\includegraphics[width=1\hsize]{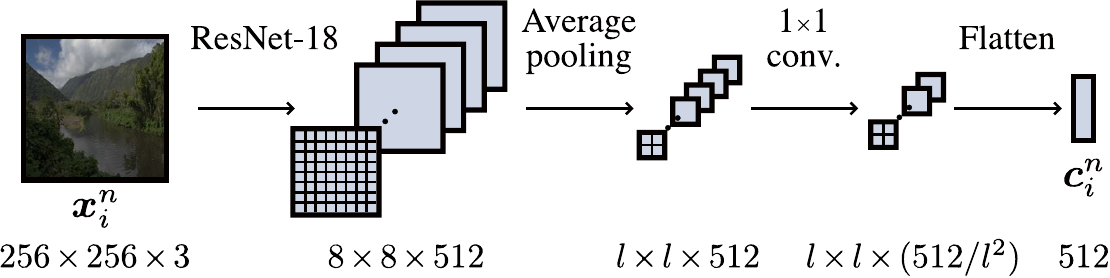}
\caption{Architecture of ${\rm f^{co}}$.}
\label{network}
\end{figure}

\vspace{2mm}
\noindent
{\bf Masked style modeling}~
For content-aware PIE, we propose masked style modeling, which is inspired by masked language modeling. In masked language modeling, each word is represented as a combination of a token embedding, a positional embedding, and a segment embedding, and the Transformer encoder predicts a masked token. Based on masked language modeling, we represent each preferred image as a combination of a style embedding and a content embedding. By replacing the token and positional embeddings with the style and content embeddings, respectively, a style embedding for an unseen image can be considered as a masked style embedding. Because the Transformer encoder can predict the masked token considering the contexts in masked language modeling, the Transformer encoder can also predict the masked style embedding considering the contents of the images. The segment embedding in masked language modeling is not used in our method because it is needed when processing a sequence of sentences.

To achieve masked style modeling, we must train four networks: a style embedding network ${\rm f^{st}}$, a content embedding network ${\rm f^{co}}$, a Transformer encoder ${\rm f^{tr}}$, and a stylized enhancer ${\rm f^{en}}$. The style embedding network and the content embedding network are needed to generate style embeddings and content embeddings from images, respectively, and the stylized enhancer is needed to enhance images based on the predicted style embeddings. In masked language modeling, all embeddings and the Transformer encoder are trained simultaneously, but we have to train the style embedding network in advance because our Transformer encoder predicts the masked style embedding. Our training process is divided into two steps. In the first step, the style embedding network ${\rm f^{st}}$ and the stylized enhancer ${\rm f^{en}}$ are trained, and in the second step, the content embedding network ${\rm f^{co}}$ and the Transformer encoder ${\rm f^{tr}}$ are trained. We show the overview in Figure~\ref{proposed} and present a detailed description in the following sections.

\subsection{Networks}
\subsubsection{Style Embedding Network\label{Style_Embedding_Network}}
The style embedding network ${\rm f^{st}}$ generates style embeddings from images.
We use the same architecture as PieNet ({\it i.e.}, a ResNet-18 followed by a global average pooling layer)
and we denote the style embedding ${\bm s}^n_i$ as
\begin{equation}
\label{diff} {\bm s}^n_i = {\rm f^{st}}({\bm y}^n_i) - {\rm f^{st}}({\bm x}^n_i).
\end{equation}
If we follow PieNet, it is more reasonable to extract ${\bm s}^n_i$ from only ${\bm y}^n_i$ as ${\bm s}^n_i = {\rm f^{st}}({\bm y}^n_i)$. In this case, however, ${\bm s}^n_i$ contains highly complex information because the color distribution of ${\bm y}^n_i$ is directly embedded into ${\bm s}^n_i$, which makes it difficult for the Transformer encoder to predict ${\bm s}^n_i$. We focus on the fact that previous enhancement methods~\cite{moran2020deeplpf,song2021starenhancer} do not predict the enhanced result directly but the residual between the input and the enhanced result because the residual is simpler than the enhanced result. Based on this fact,
we define ${\bm s}^n_i$ as Eq.~(\ref{diff}) to make the information of ${\bm s}^n_i$ simpler.
See Table~\ref{result_style_embedding} for more details.

\subsubsection{Content Embedding Network}
The content embedding network ${\rm f^{co}}$ generates a style embedding ${\bm c}^n_i$ from an original image ${\bm x}^n_i$ as
\begin{equation}
{\bm c}^n_i = {\rm f^{co}}({\bm x}^n_i).
\end{equation}
Because human retouching depends on the layout of an image, ${\bm c}^n_i$ should have spatial information. To allow ${\bm c}^n_i$ to have spatial information, we use the architecture of ${\rm f^{co}}$ in Figure~\ref{network}, where ${\bm x}^n_i$, which is resized to $256\times256$ px, is converted to a feature map of $l\times l\times (512/l^2)$, and the flattened feature is used as ${\bm c}^n_i$.

\subsubsection{Transformer Encoder}
The Transformer encoder ${\rm f^{tr}}$ receives the style embeddings and content embeddings and predicts the style embedding of an unseen input.
When training ${\rm f^{tr}}$, we select $(I_{\rm train}+1)$ preferred images of a known user and mask the style embedding of the $(I_{\rm train}+1)$th image, and the networks are trained to predict the masked style embedding from the other $I_{\rm train}$ images. We denote the input for ${\rm f^{tr}}$ as ${\bm A}$,
\begin{equation}\begin{split}
  {\bm A} = [{\bm c}^n_1\oplus{\bm s}^n_1,..., {\bm c}^n_{I_{\rm train}}\oplus{\bm s}^n_{I_{\rm train}}, {\bm c}^n_{I_{\rm train}+1}\oplus{\bm s}_{\rm masked}],
\end{split}\end{equation}
where $\oplus$ means the concatenation, and ${\bm s}_{\rm masked}$ is a learnable embedding.
The Transformer encoder outputs a matrix of the same size as ${\bm A}$,
and we apply a fully connected layer to the last column vector to predict ${\bm s}^n_{I_{\rm train}+1}$.
We denote the output as ${\rm f^{tr}}({\bm A})$.

\subsubsection{Stylized Enhancer}
The stylized enhancer ${\rm f^{en}}$ enhances an image ${\bm x}^n_i$ based on a style embedding ${\bm s}^{n'}_{i'}$.
We use the same architecture as PieNet, {\it i.e.}, a modified U-net where style embeddings can be inserted into the skip connections,
and denote the output as ${\rm f^{en}}({\bm x}^n_i, {\bm s}^{n'}_{i'})$.

\subsection{Training and Testing}
\subsubsection{First Training Step}
In the first training step, we train the style embedding network ${\rm f^{st}}$ and the stylized enhancer ${\rm f^{en}}$.
We change the training procedure from PieNet because the loss in Eq.~(\ref{pienetloss1}) does not consider a case
where preferred styles depend on the contents of images.
When the preferred styles depend on the contents, the styles of ${\bm y}^n_i$ and ${\bm y}^n_{i'}$ may be different;
however, both ${\rm f^{st}}({\bm y}^n_i)$ and ${\rm f^{st}}({\bm y}^n_{i'})$ always get close to ${\bm v}^n$.
To avoid this problem, we conclude that each style embedding should be independent.
To make each style embedding independent,
we omit the preference vector ${\bm v}^n$ and train ${\rm f^{st}}$ and ${\rm f^{en}}$ simultaneously using the following loss,
\begin{equation}
L_1 = {\rm loss^{pienet}}({\bm y}^n_i, {\rm f^{en}}({\bm x}^n_i, {\bm s}^n_i)).
\end{equation}
By training the networks using this loss, the styles of each image are independently embedded into ${\bm s}^n_i$.

\subsubsection{Second Training Step}
In the first training step, we train the content embedding network ${\rm f^{co}}$ and the Transformer encoder ${\rm f^{tr}}$.
When training these networks, we select $(I_{\rm train}+1)$ preferred images of a known user and mask the style embedding of the $(I_{\rm train}+1)$th image,
and the networks are trained to predict the masked style embedding from the other $I_{\rm train}$ images.
${\rm f^{co}}$ and ${\rm f^{tr}}$ are trained at the same time by minimizing the following loss,
\begin{equation}\label{l2}
  L_2 = {\rm MAE} ({\bm s}^n_{I_{\rm train}+1}, {\rm f^{tr}}({\bm A})),
\end{equation}
where ${\rm MAE}()$ calculates a mean absolute error.
${\mathcal S}^n=\{{\bm x}^n_i, {\bm y}^n_i\}_{i=1}^{I_n}$ is randomly reordered at each iteration.

\subsubsection{Testing}
When testing, based on a new user's preferred image set ${\mathcal S}^{\rm new}=\{{\bm x}^{\rm new}_i, {\bm y}^{\rm new}_i\}_{i=1}^{I_{\rm new}}$, $\{{\bm c}^{\rm new}_i\}_{i=1}^{I_{\rm new}}$ and $\{{\bm s}^{\rm new}_i\}_{i=1}^{I_{\rm new}}$
are obtained. In addition, ${\bm c}_{\rm unseen}$ is obtained from an unseen image ${\bm x}_{\rm unseen}$. An input for ${\rm f^{tr}}$ is represented as
\begin{equation}\begin{split}
  {\bm A^{\rm new}} = [{\bm c}^{\rm new}_1\oplus{\bm s}^{\rm new}_1,...,\,&{\bm c}^{\rm new}_{I_{\rm new}}\oplus{\bm s}^{\rm new}_{I_{\rm new}}, \\ &{\bm c}_{\rm unseen}\oplus{\bm s}_{\rm masked}],
\end{split}\end{equation}
and the enhanced result ${\bm y}^{\rm new}_{\rm unseen}$ is obtained as
\begin{equation}\begin{split}
  {\bm y}^{\rm new}_{\rm unseen} = {\rm f^{en}}\bigl({\bm x}_{\rm unseen}, {\rm f^{tr}}({\bm A^{\rm new}})\bigr).
\end{split}\end{equation}
Fine-tuning is not required in this process.
Because the Transformer encoder can receive variable length inputs,
$I_{\rm new}$ can be different for each user
and different from $I_{\rm train}$.

\section{Training Scheme}
To train our PIE model, we need a large number of user-preferred images.
In previous datasets, FiveK~\cite{bychkovsky2011learning} and PPR10K~\cite{liang2021ppr10k}, images are retouched by multiple experts, but the numbers of experts are only five and three, respectively.
In a training scheme of PieNet~\cite{kim2020pienet}, to increase the number of users,
rule-based enhancement methods ({\it i.e.}, conventional image enhancement methods and presets in Adobe Lightroom)
are applied to original images in FiveK; these rule-based methods were used as pseudo-users.
However, because the rule-based methods apply the same effects to all images and do not consider the contents, they are not suitable for content-aware PIE.

To allow our model to consider the contents of images,
we propose a novel training scheme where we use multiple real users' preferred images;
to this end, we use images on Flickr.
Flickr users retouch images according to their preferences considering the contents,
which makes Flickr images suitable for content-aware PIE.
We download 100 images each from 1,000 users,
and a total of 100,000 images are collected.

\begin{figure}[t]
\centering
\includegraphics[width=1\hsize]{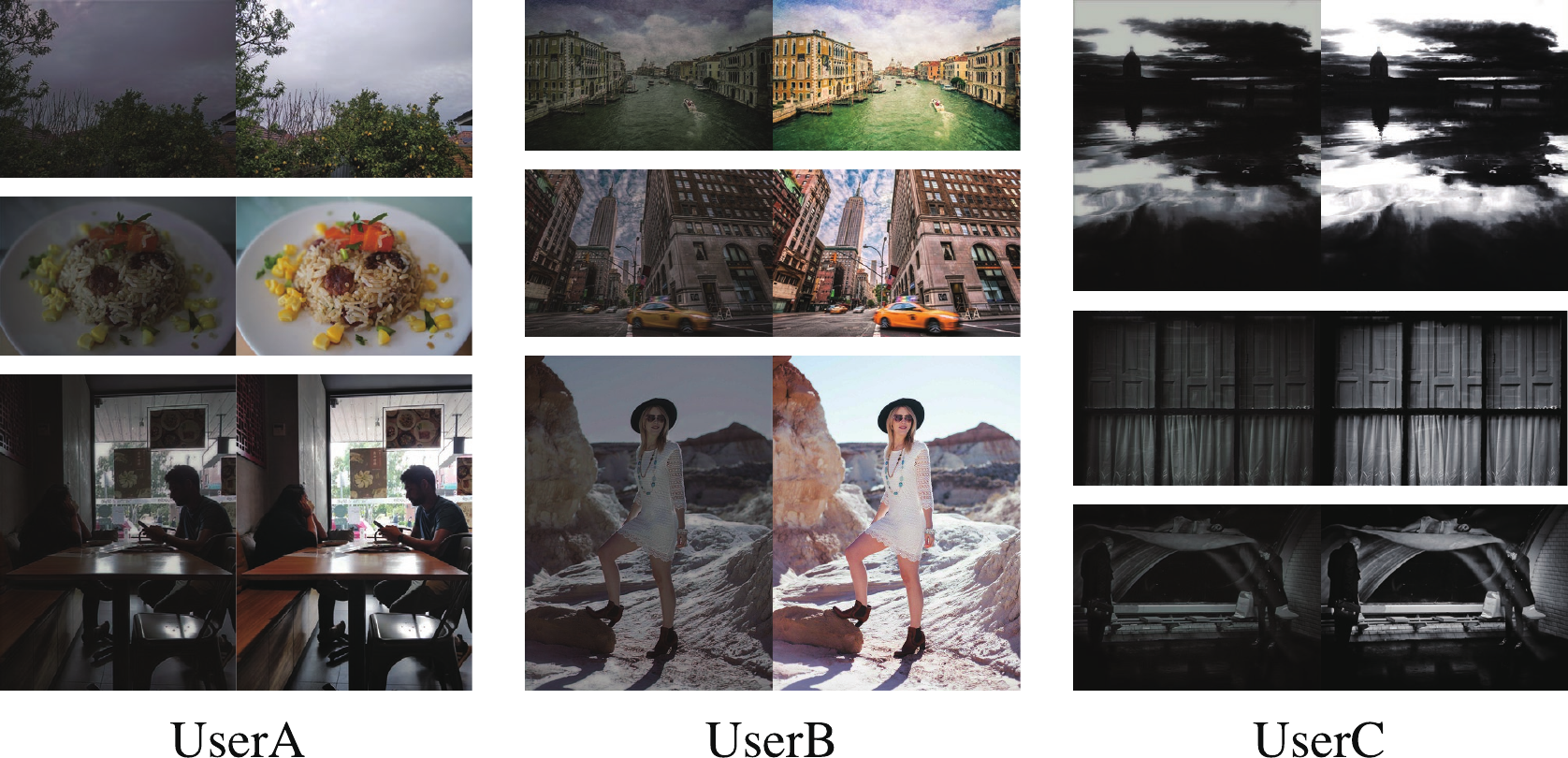}
\caption{Examples of images used in our training scheme.}
\label{dataset}
\end{figure}

To train enhancement models, original and retouched image pairs are needed;
however, only retouched images are uploaded on Flickr.
To create pseudo original and retouched image pairs, we train a degrading model.
We apply rule-based enhancement methods (10 conventional image enhancement
methods~\cite{wang2017contrast,reza2004realization,arici2009histogram,cai2017joint,lee2013contrast,guo2016lime,aubry2014fast,wang2013naturalness,fu2015probabilistic,fu2016weighted} and 15 presets in Adobe Lightroom) to original images in FiveK and
train an enhancement model in the opposite way as usual:
the model is trained to transfer the enhanced images to the original images;
the trained model can be used as a degrading model.
We use DSN~\cite{zhao2021deep} as the degrading model.
By degrading the retouched images using this degrading model,
we obtain pseudo original and retouched image pairs.
Examples of original and retouched image pairs are presented in Figure~\ref{dataset},
which shows that Flickr users have different preferences, and the degrading model properly degrades the retouched images.
By training our model using these images,
our model can consider the contents of images.

\begin{table*}[t]
  \centering
\caption{Quantitative comparisons on FiveK and PPR10K.
\dag~means fine-tuned image enhancement models.
Bold fonts and underlines mean the best and second-best scores, respectively.}
	{\tabcolsep=1.4mm
	\begin{tabular}{lcccccccccccc}\toprule[0.4mm]
\multirow{2}{*}{\textbf{Method}} &\multicolumn{3}{c}{\textbf{PSNR}~$\uparrow$} & &\multicolumn{3}{c}{\textbf{SSIM}~$\uparrow$} & &\multicolumn{3}{c}{$\bm{\Delta E_{ab}}\downarrow$}\\
\cmidrule{2-4}\cmidrule{6-8}\cmidrule{10-12}
&$\bm{I_{\rm new}=20}$ &$\bm{50}$ &$\bm{100}$ & &$\bm{20}$ &$\bm{50}$ &$\bm{100}$ &  &$\bm{20}$ &$\bm{50}$ &$\bm{100}$ \\
\toprule[0.4mm]
\multicolumn{12}{c}{\textbf{Results on FiveK}} \\ \midrule
DeepUPE$^{\dag}$ &16.51±0.18 &16.80±0.15 &16.80±0.08 & &0.788±0.004 &0.795±0.003 &0.795±0.002 & &19.00±0.30 &18.53±0.24 &18.53±0.14 \vspace{0.5mm}\\
DeepLPF$^{\dag}$ &20.14±0.16 &20.29±0.13 &20.51±0.12 & &0.858±0.002 &0.860±0.003 &0.863±0.001 & &12.61±0.24 &12.32±0.15 &11.98±0.11 \vspace{0.5mm}\\
CSRNet$^{\dag}$ &21.47±0.34 &21.86±0.22 &22.15±0.27 & &0.873±0.007 &0.878±0.004 &0.881±0.005 & &13.38±2.10 &12.04±0.79 &11.91±1.28 \vspace{0.5mm}\\
3DLUT$^{\dag}$ &21.94±0.20 &\underline{22.16±0.17} &22.30±0.13 & &\underline{0.884±0.003} &\underline{0.886±0.004} &0.889±0.003 & &11.54±0.20 &11.33±0.22 &11.14±0.17 \vspace{0.5mm}\\
Exposure Correction$^{\dag}$ &21.31±0.32 &21.60±0.21 &21.94±0.16 & &0.859±0.003 &0.862±0.002 &0.868±0.003 & &12.19±0.37 &11.89±0.29 &11.55±0.16 \vspace{0.5mm}\\
DSN$^{\dag}$ &20.54±0.34 &20.76±0.38 &21.14±0.22 & &0.849±0.007 &0.856±0.008 &0.860±0.006 & &14.86±0.74 &14.29±0.74 &13.79±0.65 \vspace{0.5mm}\\
STAR$^{\dag}$ &21.51±0.39 &21.45±0.34 &21.73±0.20 & &0.877±0.008 &0.877±0.008 &0.883±0.003 & &11.96±0.47 &11.99±0.45 &11.63±0.24 \vspace{0.5mm}\\
RCTNet$^{\dag}$ &21.75±0.32 &21.87±0.39 &\underline{22.41±0.29} & &0.880±0.005 &0.881±0.006 &\underline{0.890±0.004} & &11.62±0.32 &11.43±0.41 &\underline{10.83±0.29} \vspace{0.5mm}\\
AdaInt$^{\dag}$ &\underline{22.06±0.20} &22.12±0.21 &22.29±0.22 & &0.882±0.001 &0.884±0.002 &0.888±0.002 & &\underline{11.30±0.27} &\underline{11.25±0.25} &11.05±0.22 \vspace{0.5mm}\\
SepLUT$^{\dag}$ &21.68±0.15 &22.00±0.15 &22.39±0.18 & &0.876±0.002 &0.880±0.003 &0.886±0.003 & &11.79±0.20 &11.45±0.15 &11.05±0.19 \vspace{0.5mm}\\
NeurOp$^{\dag}$ &20.79±0.19 &21.26±0.09 &21.82±0.32 & &0.863±0.005 &0.872±0.003 &0.882±0.005 & &13.03±0.27 &12.51±0.12 &11.95±0.32 \vspace{0.5mm}\\
SpliNet &18.74±0.34 &18.94±0.41 &19.94±0.15 & &0.819±0.007 &0.824±0.007 &0.840±0.002 & &15.73±0.65 &15.35±0.58 &13.87±0.19 \vspace{0.5mm}\\
PieNet &20.52±0.09 &20.47±0.09 &20.54±0.08 & &0.850±0.003 &0.849±0.003 &0.851±0.003 & &13.55±0.16 &13.65±0.14 &13.56±0.11 \vspace{0.5mm}\\
StarEnhancer &19.87±0.22 &19.68±0.22 &19.68±0.20 & &0.839±0.008 &0.832±0.008 &0.833±0.007 & &14.66±0.31 &14.95±0.32 &14.94±0.30 \vspace{0.5mm}\\
\cellcolor[HTML]{efefef}Ours &\cellcolor[HTML]{efefef}\textbf{22.98±0.18} &\cellcolor[HTML]{efefef}\textbf{23.13±0.08} &\cellcolor[HTML]{efefef}\textbf{23.18±0.05} &\cellcolor[HTML]{efefef}\textbf{} &\cellcolor[HTML]{efefef}\textbf{0.897±0.002} &\cellcolor[HTML]{efefef}\textbf{0.897±0.002} &\cellcolor[HTML]{efefef}\textbf{0.898±0.002} &\cellcolor[HTML]{efefef}\textbf{} &\cellcolor[HTML]{efefef}\textbf{10.40±0.25} &\cellcolor[HTML]{efefef}\textbf{10.21±0.10} &\cellcolor[HTML]{efefef}\textbf{10.13±0.05} \vspace{0.5mm}\\
\toprule[0.4mm]
\multicolumn{12}{c}{\textbf{Results on PPR10K}} \\
\midrule
DeepUPE$^{\dag}$ &17.22±1.36 &17.72±0.72 &18.11±0.34 & &0.816±0.028 &0.837±0.016 &0.852±0.009 & &17.49±2.25 &16.44±1.14 &15.85±0.53 \vspace{0.5mm}\\
DeepLPF$^{\dag}$ &19.90±0.51 &20.50±0.27 &20.86±0.27 & &0.881±0.007 &0.890±0.003 &0.893±0.002 & &12.27±0.49 &11.52±0.30 &11.15±0.28 \vspace{0.5mm}\\
CSRNet$^{\dag}$ &20.60±0.28 &21.15±0.26 &21.70±0.10 & &0.891±0.003 &0.897±0.004 &0.905±0.002 & &13.08±0.33 &12.26±0.35 &11.55±0.12 \vspace{0.5mm}\\
3DLUT$^{\dag}$ &20.95±0.14 &20.89±0.21 &21.48±0.15 & &0.888±0.002 &0.895±0.003 &0.903±0.004 & &12.40±0.16 &11.99±0.19 &11.43±0.20 \vspace{0.5mm}\\
Exposure Correction$^{\dag}$ &21.44±0.14 &21.24±0.29 &21.24±0.10 & &0.876±0.003 &0.875±0.003 &0.875±0.001 & &11.73±0.20 &11.88±0.37 &11.85±0.17 \vspace{0.5mm}\\
DSN$^{\dag}$ &20.07±0.46 &20.36±0.52 &20.77±0.30 & &0.876±0.007 &0.884±0.006 &0.890±0.004 & &14.92±1.17 &14.81±1.58 &13.23±0.93 \vspace{0.5mm}\\
STAR$^{\dag}$ &20.86±0.35 &20.92±0.16 &21.16±0.19 & &0.900±0.004 &0.904±0.003 &\underline{0.910±0.002} & &12.44±0.39 &12.34±0.20 &11.73±0.24 \vspace{0.5mm}\\
RCTNet$^{\dag}$ &\underline{21.78±0.34} &\underline{21.94±0.30} &\underline{22.16±0.22} & &\underline{0.906±0.005} &\underline{0.911±0.003} &0.909±0.009 & &\underline{11.34±0.44} &\underline{10.96±0.28} &\underline{10.62±0.25} \vspace{0.5mm}\\
AdaInt$^{\dag}$ &21.29±0.44 &21.26±0.55 &21.58±0.39 & &0.897±0.004 &0.902±0.005 &0.909±0.003 & &11.78±0.44 &11.64±0.53 &11.15±0.38 \vspace{0.5mm}\\
SepLUT$^{\dag}$ &20.76±0.27 &20.96±0.33 &21.36±0.20 & &0.886±0.005 &0.892±0.005 &0.902±0.004 & &12.31±0.41 &11.96±0.36 &11.38±0.26 \vspace{0.5mm}\\
NeurOp$^{\dag}$ &21.31±0.19 &21.63±0.33 &22.07±0.18 & &0.891±0.005 &0.901±0.004 &0.909±0.003 & &12.26±0.30 &11.79±0.52 &11.28±0.25 \vspace{0.5mm}\\
SpliNet &19.15±1.01 &19.58±0.98 &20.34±0.22 & &0.849±0.009 &0.851±0.013 &0.865±0.006 & &14.70±1.41 &14.41±1.53 &12.84±0.18 \vspace{0.5mm}\\
PieNet &19.94±0.07 &19.95±0.05 &19.97±0.05 & &0.821±0.006 &0.823±0.004 &0.824±0.004 & &14.40±0.16 &14.36±0.13 &14.32±0.12 \vspace{0.5mm}\\
StarEnhancer &21.17±0.06 &21.21±0.05 &21.22±0.04 & &0.883±0.004 &0.885±0.002 &0.885±0.001 & &12.33±0.09 &12.29±0.06 &12.29±0.05 \vspace{0.5mm}\\
\cellcolor[HTML]{efefef}Ours &\cellcolor[HTML]{efefef}\textbf{22.78±0.16} &\cellcolor[HTML]{efefef}\textbf{22.88±0.19} &\cellcolor[HTML]{efefef}\textbf{22.97±0.13} &\cellcolor[HTML]{efefef}\textbf{} &\cellcolor[HTML]{efefef}\textbf{0.918±0.002} &\cellcolor[HTML]{efefef}\textbf{0.920±0.002} &\cellcolor[HTML]{efefef}\textbf{0.921±0.001} &\cellcolor[HTML]{efefef}\textbf{} &\cellcolor[HTML]{efefef}\textbf{10.09±0.24} &\cellcolor[HTML]{efefef}\textbf{9.95±0.18} &\cellcolor[HTML]{efefef}\textbf{9.83±0.11} \vspace{0.5mm}\\
\bottomrule[0.4mm]
\end{tabular}
}
\label{result1}
\end{table*}

\begin{figure*}[t]
  \centering
  \includegraphics[width=1\hsize]{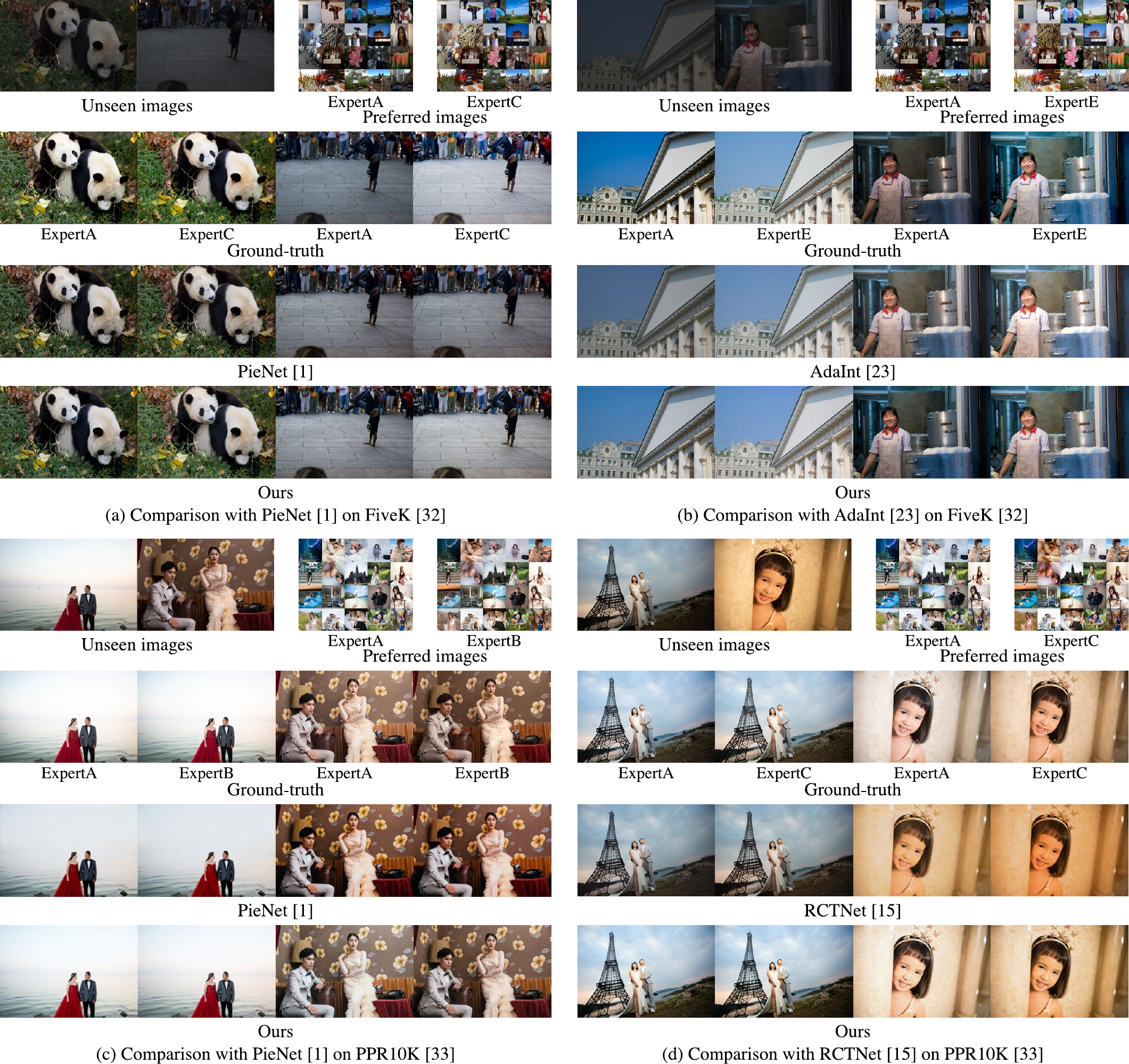}
\caption{Qualitative comparisons.
PieNet~\cite{kim2020pienet} is the baseline PIE method,
and AdaInt~\cite{yang2022adaint} and RCTNet~\cite{kim2021representative} are fine-tuned image enhancement models
which achieve the second-best performance in each dataset.
In (a),
the two experts have similar preferences for the left image but have different preferences for the right image,
and PieNet cannot reproduce the image-adaptive preferences because the contents are not considered.
The proposed method can efficiently predict the preferences considering the contents.
In (b), AdaInt generates worse results than the proposed method
because AdaInt is not suited to be fine-tuned with a limited number of preferred images.
Similarly, in (c), the two experts have similar preferences for the left image but have different preferences for the right image, and the proposed method can efficiently predict the preferences considering the contents;
in (d), our method generates better results than RCTNet.
}
\label{results_vis}
\end{figure*}

\section{Experiments}
\subsection{Datasets and Implementation}
In addition to the Flickr images, we use two datasets: {\bf FiveK}~\cite{bychkovsky2011learning} and {\bf PPR10K}~\cite{liang2021ppr10k}. We use 900 users of the Flickr users as known users for training and the rest 100 users for validation. In addition, we apply 25 rule-based methods (10 conventional image enhancement
methods~\cite{wang2017contrast,reza2004realization,arici2009histogram,cai2017joint,lee2013contrast,guo2016lime,aubry2014fast,wang2013naturalness,fu2015probabilistic,fu2016weighted} and 15 presets in Adobe Lightroom) to 4,500 images out of the 5,000 images in FiveK and use these rule-based methods as known users for training. When testing, we use five experts in FiveK and three experts in PPR10K as new users. When testing with FiveK, we randomly select ${\mathcal S}^{\rm new}$
from the 4,500 images and use the remaining 500 as ${\bm x}_{\rm unseen}$; when testing with PPR10K, we randomly select ${\mathcal S}^{\rm new}$ from 8,875 images and use the remaining 2,286 images as ${\bm x}_{\rm unseen}$. We perform evaluations setting ${I_{\rm new}}$ to $\{20, 50, 100\}$. Because the performance depends on how ${\mathcal S}^{\rm new}$ is selected, we sample ${\mathcal S}^{\rm new}$ 10 times and present the mean and standard deviation of evaluation metrics. Following \cite{yang2022adaint}, we use PSNR, SSIM, and $\Delta E_{ab}$ as evaluation metrics; higher PSNR and SSIM scores mean better results, and a lower $\Delta E_{ab}$ score means better results.
We train the networks using Adam~\cite{kingma2014adam} with a learning rate of $1.0 \times 10^{-4}$
for 40 and 30 epochs, and the batch size is set to 64 and 32 in the first and second training steps, respectively.
$I_{\rm train}$ is set to 10.
All images are resized to $256 \times 256$ px when input to the content and style embedding networks and resized to $512 \times 512$ px when input to the stylized enhancer.

\subsection{Comparisons with Previous Methods}
We use 11 recent general image enhancement models: DeepUPE~\cite{wang2019underexposed},  DeepLPF~\cite{moran2020deeplpf}, CSRNet~\cite{he2020conditional}, 3DLUT~\cite{zeng2022learning}, Exposure Correction~\cite{afifi2021learning}, DSN~\cite{zhao2021deep}, STAR~\cite{zhang2021star}, RCTNet~\cite{kim2021representative}, AdaInt~\cite{yang2022adaint}, SepLUT~\cite{yang2022seplut}, and NeurOp~\cite{wang2022neural}. We pre-train these models using all ${\bm x}^n_i$ and ${\bm y}^n_i$ and fine-tune them using ${\mathcal S}^{\rm new}$. We fine-tune each model for 10 epochs using 90\% of ${\mathcal S}^{\rm new}$ for training and the rest 10\% as validation, and model weights with the best validation scores are selected. We use SpliNet~\cite{bianco2020personalized}, PieNet~\cite{kim2020pienet}, and StarEnhancer~\cite{song2021starenhancer} as PIE methods.

The results of the quantitative comparisons are shown in Table~\ref{result1}. The performances of the previous PIE methods are relatively low because they do not consider the contents. Recent general enhancement models outperform the PIE methods although they require fine-tuning. Compared with these methods, the proposed method achieves the best performance even though it does not require fine-tuning. In addition, the performance of the proposed method is improved by increasing $I_{\rm new}$. In Figures~\ref{results_vis}(a) and (b), we qualitatively compare the proposed method with the baseline method ({\it i.e.,} PieNet) and the second-best method ({\it i.e.,} AdaInt) with $I_{\rm new}=20$ on FiveK. In comparison with PieNet, the two experts have similar preferences for the left image but have different preferences for the right image, and PieNet cannot reproduce the image-adaptive preferences because the contents are not considered. Compared with PieNet, the proposed method can efficiently predict the preferences considering the contents.
AdaInt is not suited to be fine-tuned with a limited number of preferred images
because of a large number of model parameters;
therefore, AdaInt generates worse results than the proposed method.
We show the qualitative comparisons on PPR10K in Figures~\ref{results_vis}(c) and (d).
In Figure~\ref{results_vis}(c),
the two experts have similar preferences for the left image but have different preferences for the right
image, and the proposed method can efficiently predict the preferences considering the contents;
in Figure~\ref{results_vis}(d),
our method generates better results than RCTNet because RCTNet is not suited to be fine-tuned with a limited number of preferred images.

\subsection{Visualization of Content-awareness}
\begin{figure*}[t]
\centering
\includegraphics[width=1\hsize]{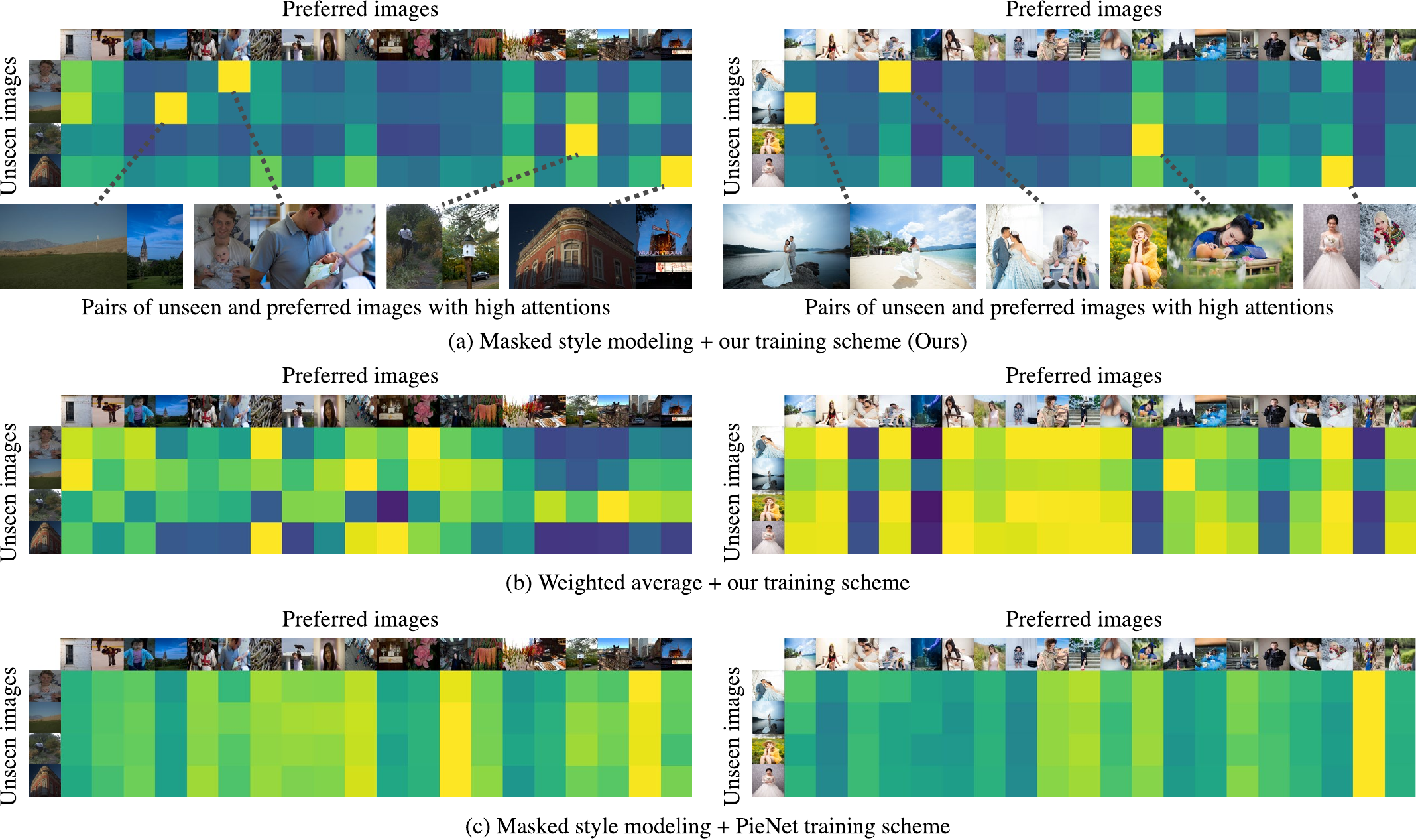}
\caption{Attentions between unseen and preferred images.
In (a), the attentions between unseen and preferred images with similar
contents are high, which indicates that our Transformer is able to consider the contents.
In (b), we replace masked style modeling with weighted average,
where the attentions are independent of the contents.
Because the weighted average is a simple attention module,
it cannot capture the complex relationships of the contents.
In (c), we train our masked style modeling using PieNet training scheme, where we use only
images enhanced by rule-based methods.
The attentions are independent of unseen images, which means that our Transformer
encoder cannot consider the contents.
Therefore, our training scheme is essential for content-aware PIE.
}
\label{attention}
\end{figure*}

To demonstrate that our Transformer encoder can consider the contents, we visualize the attentions between unseen and preferred images using attention flow~\cite{abnar2020quantifying} in Figure~\ref{attention}(a). The attentions between unseen and preferred images with similar contents are high, which indicates that our Transformer is able to consider the contents.

\begin{figure*}[t]
  \centering
  \includegraphics[width=1\hsize]{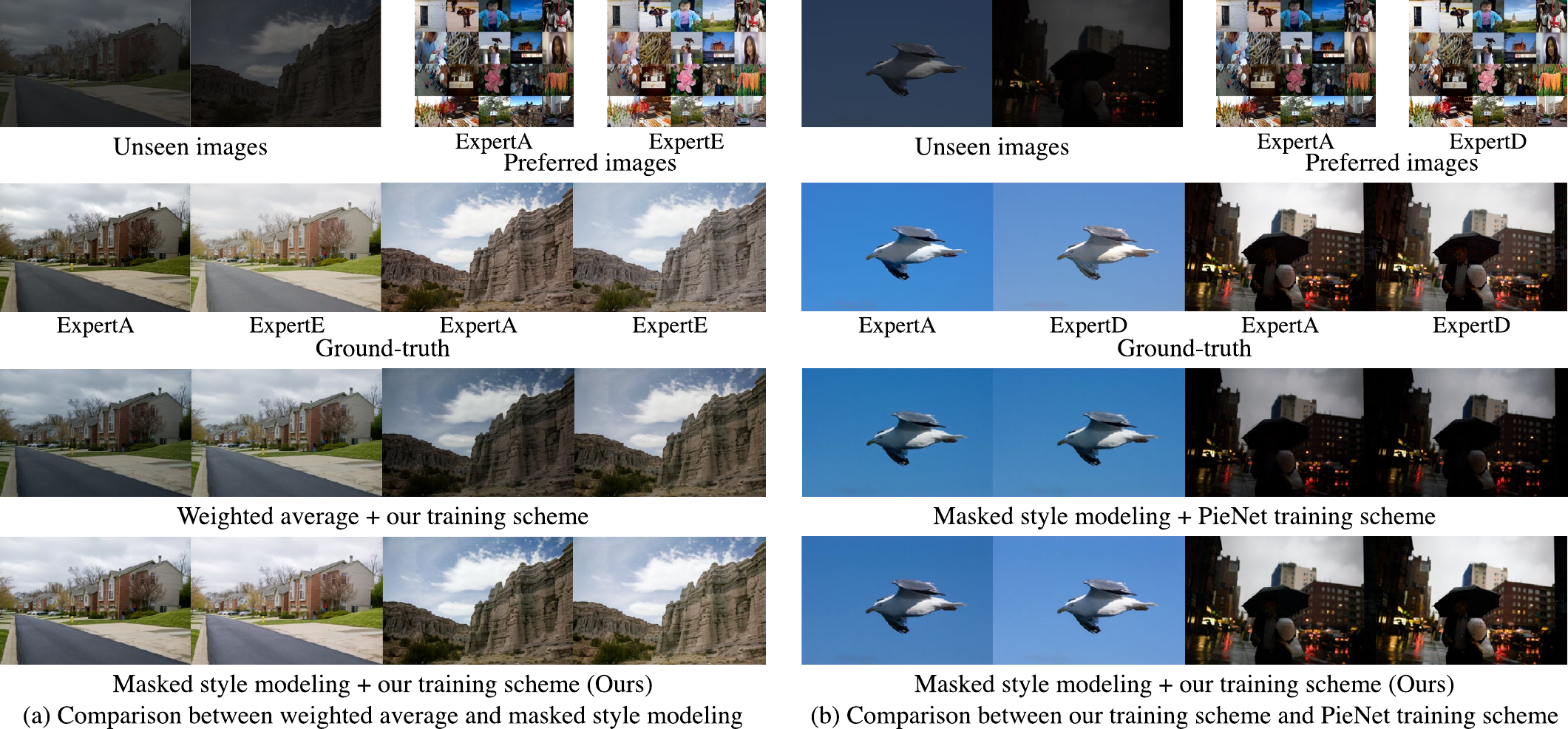}
\caption{Qualitative comparisons when changing the attention module and the training scheme.
In (a), we replace masked style modeling with weighted average,
and our masked style modeling generates better results.
In (b), we replace our training scheme with PieNet training scheme,
and masked style modeling trained with our training scheme generates better results.
}
\label{results_vis_ablation}
\end{figure*}

\subsection{Attention Module\label{attention_module}}
We propose masked style modeling to achieve content-aware PIE.
To show that our masked style modeling can consider the contents efficiently,
we replace the Transformer encoder with a simple attention module, weighted average.
In content-aware PIE, we have to apply similar styles to images which have similar contents.
To achieve such content-aware PIE simply,
we calculate the similarity between the content embedding of an unseen image
and the content embeddings of preferred images;
then, we multiply the style embeddings of all preferred images by the similarity.
We take an average of the multiplied style embeddings to predict the style embedding for the unseen image.
The training loss in the second training step (\ref{l2}) is replaced as
\begin{equation}
  L_2 = {\rm MAE} ({\bm s}^n_{I_{\rm train}+1}, 1/{W^n}\sum\nolimits_{i=1}^{I_{\rm train}}w^n_i {\bm s}^n_i),
\end{equation}
where
\begin{equation}
  w^n_i = {\rm sim}({\bm c}^n_i, {\bm c}^n_{I_{\rm train}+1}),~~W^n = \sum\nolimits_{i=1}^{I_{\rm train}}w^n_i,
\end{equation}
and ${\rm sim}()$ calculates cosine similarity.
When testing, the enhanced result ${\bm y}^{\rm new}_{\rm unseen}$ is obtained as
\begin{equation}\begin{split}
  {\bm y}^{\rm new}_{\rm unseen} = {\rm f^{en}}\bigl({\bm x}_{\rm unseen}, 1/{W^{\rm new}}\sum\nolimits_{i=1}^{I_{\rm new}}w^{\rm new}_i {\bm s}^{\rm new}_i\bigr).
\end{split}\end{equation}
where
\begin{equation}
  w^{\rm new}_i = {\rm sim}({\bm c}^{\rm new}_i, {\bm c}_{\rm unseen}),~~W^{\rm new} = \sum\nolimits_{i=1}^{I_{\rm new}}w^{\rm new}_i.
\end{equation}

We show the results when replacing the Transformer encoder with the weighted average in Table~\ref{result_scheme_attention}
and qualitative comparisons in Figure~\ref{results_vis_ablation}.
Our masked style modeling achieves better performance than the weighted average.
$w^{\rm new}_i$ can be seen as attention, and we visualize $w^{\rm new}_i$ in Figure~\ref{attention}(b).
Although the weighted average is trained using our training scheme,
the attentions are independent of the contents.
These results mean that the complex relationships of the contents cannot be captured by this simple attention module,
and our masked style modeling is important for content-aware PIE.

\begin{table}[t]
  \centering
\caption{Results when changing the attention module and the training scheme.}
	{\tabcolsep=1.4mm
	\begin{tabular}{lccc}\toprule[0.4mm]
\multirow{2}{*}{\textbf{Method}} &\multicolumn{3}{c}{\textbf{PSNR}~$\uparrow$} \\
\cmidrule{2-4}
&$\bm{I_{\rm new}=20}$ &$\bm{50}$ &$\bm{100}$  \\
\toprule[0.4mm]
\cellcolor[HTML]{efefef}{\tabcolsep=0mm\begin{tabular}{l}Masked style modeling +\\ our training scheme (Ours)\end{tabular}} &\cellcolor[HTML]{efefef}\textbf{22.98±0.18} &\cellcolor[HTML]{efefef}\textbf{23.13±0.08} &\cellcolor[HTML]{efefef}\textbf{23.18±0.05} \vspace{1.5mm}\\
{\tabcolsep=0mm\begin{tabular}{l}Weighted average +\\ our training scheme\end{tabular}} & 22.02±0.12 & 22.15±0.08 & 22.18±0.09 \vspace{1.5mm}\\
{\tabcolsep=0mm\begin{tabular}{l}Masked style modeling +\\ PieNet training scheme\end{tabular}} &22.09±0.07 &22.17±0.05&22.24±0.04 \vspace{0mm}\\
\bottomrule[0.4mm]
\end{tabular}
}
\label{result_scheme_attention}
\end{table}

\subsection{Training Scheme}
In our training scheme, we train our model using image pairs created from Flickr images.
To demonstrate the importance of our training scheme,
we train our model using PieNet training scheme,
where we use only images enhanced by rule-based methods.
We show the result in Table~\ref{result_scheme_attention} and qualitative comparisons in Figure~\ref{results_vis_ablation}, where PieNet training scheme drops the enhancement performance. The attentions with PieNet training scheme are shown in Figure~\ref{attention}(c); the attentions are independent of unseen images, which means that our Transformer encoder trained with the rule-based enhancement methods cannot consider the contents.
Therefore, our training scheme is essential for content-aware PIE.

\subsection{Design of the Embeddings}

\subsubsection{Style Embedding}
We define ${\bm s}^n_i$ as Eq.~(\ref{diff}).
To verify that this design contributes to the performance, we replace the style embedding with ${\bm s}^n_i = {\rm f^{st}}({\bm y}^n_i)$. The result in Table~\ref{result_style_embedding} shows that our design of the style embedding is an important factor for the high performance.

\subsubsection{Content Embedding}
We design ${\bm c}^n_i$ to have spatial information. To verify that this design contributes to the performance, we conduct experiments setting $l=\{1,2,4,8\}$. The results in Table~\ref{result_content_embedding} show that the performances with $l=2$ and $l=4$ are higher than the performance with $l=1$. This indicates that the spatial information improves the performance. When $l=8$, the performance is a little worse because fewer dimensions are allocated to each spatial patch. In all other experiments, we set $l=2$.

\begin{table}[t]
  \centering
\caption{Results when using different design of the style embedding ${\bm s}^n_i$.}
	{\tabcolsep=1.4mm
	\begin{tabular}{lccc}\toprule[0.4mm]
\multirow{2}{*}{\textbf{Method}} &\multicolumn{3}{c}{\textbf{PSNR}~$\uparrow$} \\
\cmidrule{2-4}
&$\bm{I_{\rm new}=20}$ &$\bm{50}$ &$\bm{100}$  \\
\toprule[0.4mm]
\cellcolor[HTML]{efefef}{\tabcolsep=0mm\begin{tabular}{l}${\bm s}^n_i = {\rm f^{st}}({\bm y}^n_i) - {\rm f^{st}}({\bm x}^n_i)$ \\(Ours)\end{tabular}} &\cellcolor[HTML]{efefef}\textbf{22.98±0.18} &\cellcolor[HTML]{efefef}\textbf{23.13±0.08} &\cellcolor[HTML]{efefef}\textbf{23.18±0.05} \vspace{0.5mm}\\
${\bm s}^n_i = {\rm f^{st}}({\bm y}^n_i)$ &22.63±0.24 & 22.77±0.12 & 22.78±0.16 \vspace{0.5mm}\\
\bottomrule[0.4mm]
\end{tabular}
}
\label{result_style_embedding}
\end{table}

\begin{table}[t]
  \centering
\caption{Results when changing $l$ in the content embedding ${\bm c}^n_i$.}
	{\tabcolsep=1.4mm
	\begin{tabular}{lccc}\toprule[0.4mm]
\multirow{2}{*}{\textbf{Method}} &\multicolumn{3}{c}{\textbf{PSNR}~$\uparrow$} \\
\cmidrule{2-4}
&$\bm{I_{\rm new}=20}$ &$\bm{50}$ &$\bm{100}$  \\
\toprule[0.4mm]
$l=1$ & 22.84±0.17 & 22.98±0.10 & 23.02±0.05 \vspace{1mm}\\
\cellcolor[HTML]{efefef}$l=2$ (Ours) &\cellcolor[HTML]{efefef}\textbf{22.98±0.18} &\cellcolor[HTML]{efefef}\textbf{23.13±0.08} &\cellcolor[HTML]{efefef}\textbf{23.18±0.05} \vspace{1mm}\\
$l=4$ &22.89±0.23 & 23.10±0.05 & 23.17±0.05 \vspace{1mm}\\
$l=8$ &22.81±0.14 & 22.96±0.11 & 23.03±0.10 \vspace{0mm}\\
\bottomrule[0.4mm]
\end{tabular}
}
\label{result_content_embedding}
\end{table}

\subsection{Number of Preferred Images}

\subsubsection{A Small Number of Preferred Images}
To investigate the performance when using a small number of preferred images,
we evaluate our method setting ${I_{\rm new}}$ to $\{1, 2, 5, 10, 20\}$ on FiveK.
For comparison, we use three general image enhancement models with relatively good performance:
3DLUT~\cite{zeng2022learning}, RCTNet~\cite{kim2021representative}, and AdaInt~\cite{yang2022adaint}.
We show the PSNR scores in Figure~\ref{large_number}(a).
Our content-aware PIE method can apply different styles to each image considering the contents,
but when ${I_{\rm new}}=1$, only one preferred style is provided,
and our method just applies the only one style to all images.
Therefore, the performance of the proposed method is not much different from the performance of the previous methods.
When ${I_{\rm new}}\ge 2$, our method outperforms the previous methods by a large margin
because multiple preferred styles are provided,
and our method can apply different styles to each image considering the contents.

\subsubsection{A Large Number of Preferred Images}
\begin{figure*}[t]
	\centering
	\includegraphics[width=1\hsize]{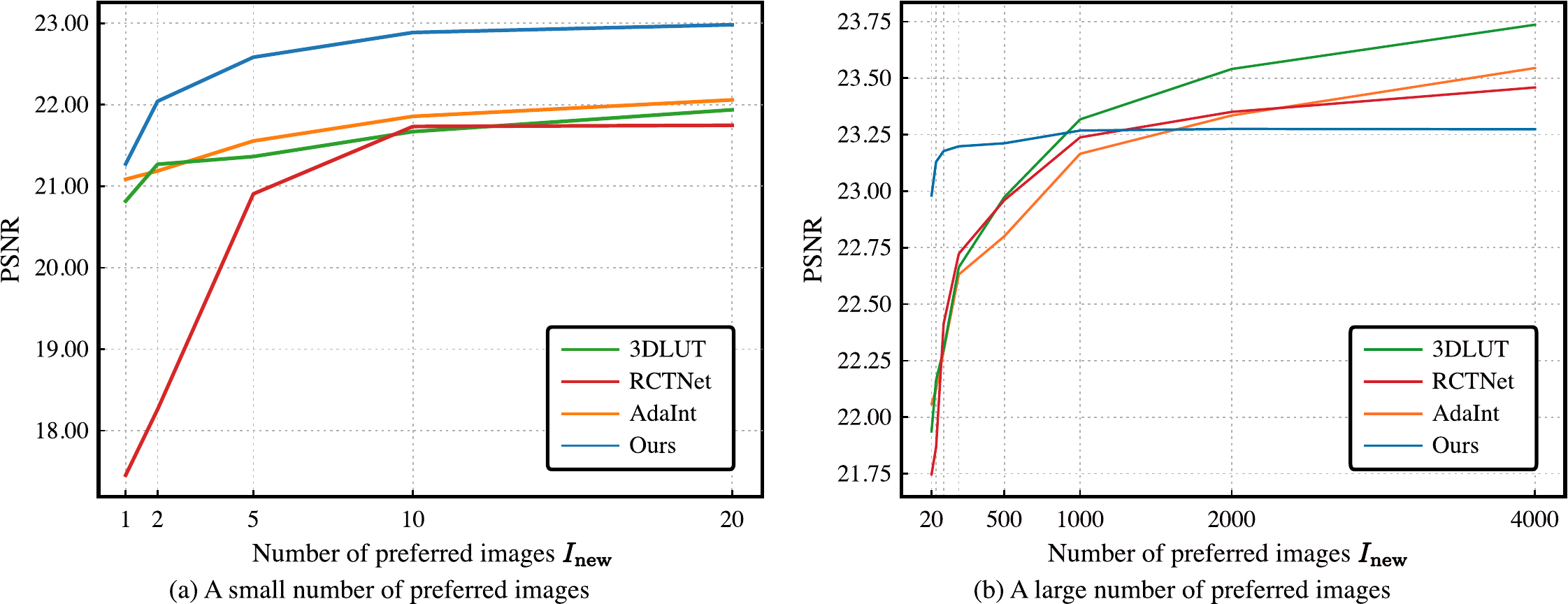}
\caption{
PSNR scores using various numbers of preferred images on FiveK. In (a), we set ${I_{\rm new}}$ to $\{1, 2, 5, 10, 20\}$,
and in (b), we set ${I_{\rm new}}$ to $\{20, 50, 100, 200, 500, 1000, 2000, 4000\}$.
}
\label{large_number}
\end{figure*}

To investigate the performance when using a large number of preferred images,
we evaluate our method setting ${I_{\rm new}}$ to $\{20, 50, 100, 200, 500, 1000, 2000, 4000\}$ on FiveK.
We show the PSNR scores in Figure~\ref{large_number}(b).
When ${I_{\rm new}}$ is greater than 1000, the performance of our method does not improve.
Because most of the contents of unseen images are included in the preferred images when ${I_{\rm new}}\ge 1000$,
new information about the user’s preference and contents of images is not provided even if ${I_{\rm new}}$ increases.
Compared to our method, the fine-tuned image enhancement models can capture more complex relationships
in a large number of images because they are specialized for each user, and the performances keep improving.
Based on this result, we can conclude that
our proposed method is the best if the number of preferred images is smaller, and
the general image enhancement models are more suitable
if a new user can provide a large number of preferred images.
In a practical scenario, it would be difficult to ask ordinary users to prepare more than 1,000 images
to show their retouch preferences, and therefore our approach would be preferred.

\begin{table}[t]\centering
	\renewcommand{\arraystretch}{1.2}
\caption{PSNR scores when excluding one of the three categories from preferred images.
In each case, we randomly select 10 preferred images from one category and 10 preferred images from another category, totally ${I_{\rm new}}=20$.}\label{category}
{\tabcolsep=1.8mm
\begin{tabular}{m{4mm}c|cccc}\toprule[0.4mm]
 & &\multicolumn{3}{c}{\textbf{Preferred images}} \\
 & &person(s) \& &nature \& &nature \& \\
 & & man-made obj. & man-made obj. & person(s) \\
 \midrule
\multirow{3}{*}{\rotatebox[origin=c]{90}{{\tabcolsep=0mm \begin{tabular}{c}\textbf{Unseen}\\ \raisebox{2mm}{\textbf{images}}\end{tabular}}}}&nature &22.42±0.17 &22.75±0.09 &22.56±0.21 \\
 &person(s) &23.58±0.14 &23.21±0.23 &23.55±0.14 \\
 &man-made obj. &23.10±0.12 &23.00±0.19 &23.03±0.13 \\
\bottomrule[0.4mm]
\end{tabular}}
\renewcommand{\arraystretch}{1}
\end{table}

\begin{table*}[t]
\centering
\caption{Comparison of the number of parameters, fine-tuning time, and inference time
where $N^{\rm new}$ is the number of new users.
\dag~means fine-tuned image enhancement models.}
{\tabcolsep=1.4mm
\begin{tabular}{lccccccccccc}\toprule[0.4mm]
\multirow{2}{*}{\textbf{Method}} & &\multirow{2}{*}{\textbf{Number of parameters}} & &\multicolumn{3}{c}{\textbf{Fine-tuning time}} & &\multicolumn{3}{c}{\textbf{Inference time}} \\
\cmidrule{5-7}\cmidrule{9-11}
& & & &$\bm{I_{\rm new}=20}$ &$\bm{50}$ &$\bm{100}$ & &$\bm{20}$ &$\bm{50}$ &$\bm{100}$ \\
\midrule
3DLUT$^{\dag}$ & & 590 K $\times N^{\rm new}$& & 5 s $\times N^{\rm new}$& 12 s $\times N^{\rm new}$& 25 s $\times N^{\rm new}$& & 0.76 ms& 0.76 ms& 0.76 ms\vspace{0.5mm}\\
RCTNet$^{\dag}$ & & 3.5 M $\times N^{\rm new}$& & 77 s $\times N^{\rm new}$& 172 s $\times N^{\rm new}$& 326 s $\times N^{\rm new}$ & & 220 ms & 220 ms & 220 ms \vspace{0.5mm}\\
AdaInt$^{\dag}$ & & 620 K $\times N^{\rm new}$& & 9 s $\times N^{\rm new}$& 15 s $\times N^{\rm new}$& 29 s $\times N^{\rm new}$ & & 1.9 ms & 1.9 ms & 1.9 ms \vspace{0.5mm}\\
PieNet & & 28 M & & 0 s & 0 s & 0 s & & 6.2 ms & 6.2 ms & 6.2 ms \vspace{0.5mm}\\
\cellcolor[HTML]{efefef}Ours &\cellcolor[HTML]{efefef} &\cellcolor[HTML]{efefef}90 M &\cellcolor[HTML]{efefef} &\cellcolor[HTML]{efefef}0 s &\cellcolor[HTML]{efefef}0 s &\cellcolor[HTML]{efefef}0 s &\cellcolor[HTML]{efefef} &\cellcolor[HTML]{efefef}13 ms &\cellcolor[HTML]{efefef}14 ms &\cellcolor[HTML]{efefef}15 ms \vspace{0.5mm}\\
\bottomrule[0.4mm]
\end{tabular}
}
\label{params}
\end{table*}

\begin{figure}[t]
	\centering
	\includegraphics[width=0.95\hsize]{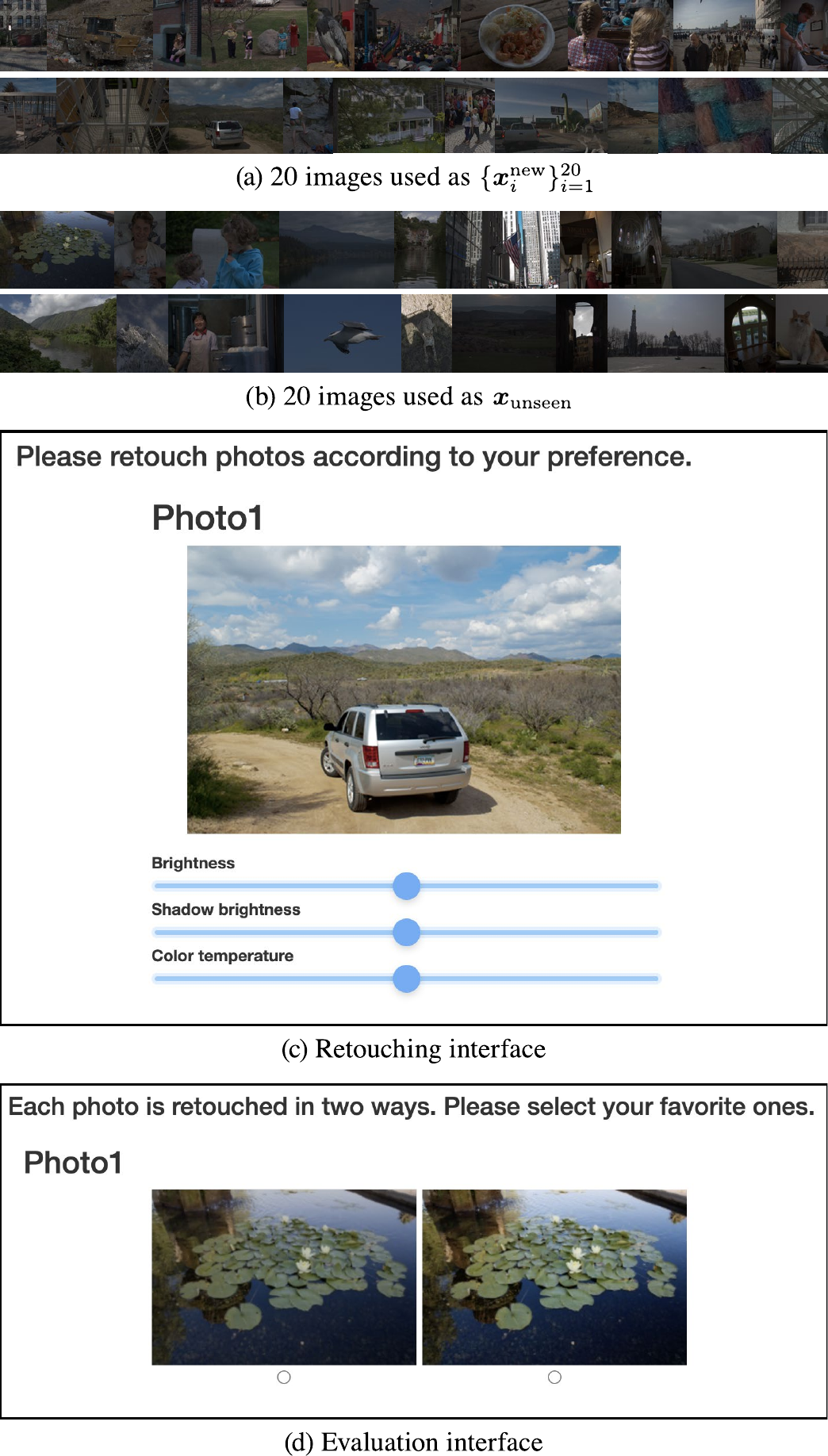}
\caption{Images and interface used in the user study.}
\label{userstudy_vis}
\end{figure}

\begin{table}[t]
\centering
\caption{Voting rates in the user study.}
{\tabcolsep=1.4mm
\begin{tabular}{lc}\toprule[0.4mm]
	\textbf{Method}&\textbf{Average voting rate}\\
	\midrule
	Ours / PieNet & \textbf{79.5\%} / 20.5\%\vspace{0.5mm}\\
	Ours / AdaInt & \textbf{70.5\%} / 29.5\%\vspace{0.5mm}\\
	Ours / AdaInt (ExpertC) & \textbf{60.5\%} / 39.5\%\vspace{0.5mm}\\
\bottomrule[0.4mm]
\end{tabular}
}
\label{userstudy}
\end{table}

\subsection{Unexpected Contents}
In a practical scenario, unseen images may have unexpected contents; {\it i.e.},
an unseen image's content may be different from any of the preferred images' contents.
To evaluate our method in such a case, we conduct an experiment using FiveK.
Each image in FiveK has one of the seven category labels about the main subject.
Among the seven categories, we use three categories: ``nature'', ``person(s)'', and ``man-made object'',
because the numbers of images with these categories are the largest.
We exclude one of the three categories from preferred images;
for example, we randomly select 10 preferred images from ``person(s)'' images and 10 preferred images from ``man-made object'' images.
We generate personalized results using such preferred images and evaluate the performance
splitting unseen images into ``nature'', ``person(s)'', and ``man-made object'' images.
We show the PSNR scores in Table~\ref{category}.
When an unseen image's content is different from any of the preferred images' contents,
the PSNR scores are relatively low.
Especially, when ``person(s)'' images are not included in the preferred images,
the performance for ``person(s)'' unseen images decreases.
These results mean that the performance of our method is degraded
if unseen images have unexpected contents.
However, the preferred images do not need to contain exactly the same content as an unseen image’s content;
they only need to contain the same broad category as an unseen image’s category.
For example, the performance for ``nature'' unseen images improves when ``nature'' images are included
in the preferred images as shown in Table~\ref{category},
where “nature” category contains a wide variety of contents such as
mountains, ponds, sky, trees, flowers, sea, rocks, etc.
Therefore, even if the number of preferred images is small,
most of the unseen images’ categories are included in the preferred images, and our method works well.

\subsection{Number of Parameters and Processing Time}
We show the number of parameters, fine-tuning time, and inference time in Table~\ref{params}.
We compare our method with relatively high-performing methods (3DLUT~\cite{zeng2022learning}, RCTNet~\cite{kim2021representative}, and AdaInt~\cite{yang2022adaint})
and the baseline method (PieNet~\cite{kim2020pienet}).
Our method and PieNet can generate personalized results for all users with a single model,
but 3DLUT, RCTNet, and AdaInt require user-specific models;
therefore, the number of parameters increases with the number of users $N^{\rm new}$.
Because our method and PieNet do not require fine-tuning,
these methods can easily generate personalized results for a large number of users;
compared to these methods, 3DLUT, RCTNet, and AdaInt require fine-tuning time for each user,
which severely limits the scalability to a large number of users.
The inference times of 3DLUT, RCTNet, and AdaInt do not depend on ${I_{\rm new}}$.
The inference time of PieNet also does not depend on ${I_{\rm new}}$ because the estimation process of a preference vector (Eq.~(\ref{average}))
is needed for each user only once.
The inference time of our method depends on ${I_{\rm new}}$ because the forwarding of the Transformer encoder
is needed for each unseen image,
but real-time inference is possible for all ${I_{\rm new}}$.

\subsection{User Study}
To evaluate the proposed method, we perform a user study. We select 20 images from FiveK for $\{{\bm x}^{\rm new}_i\}_{i=1}^{20}$. To make the contents of the 20 images as diverse as possible, we extract content embeddings from the 4,500 images and cluster them into 20 classes by using k-medoids clustering; then, the 20 images corresponding to the 20 medoids are used as $\{{\bm x}^{\rm new}_i\}_{i=1}^{20}$.
We show $\{{\bm x}^{\rm new}_i\}_{i=1}^{20}$ in Figure~\ref{userstudy_vis}(a).
For ${\bm x}_{\rm unseen}$, 20 images are randomly selected from the 500 images in FiveK,
which are shown in Figure~\ref{userstudy_vis}(b).
We hire 10 subjects.
We first present the retouching interface (Figure~\ref{userstudy_vis}(c)) to the 10 subjects and ask them to retouch $\{{\bm x}^{\rm new}_i\}_{i=1}^{20}$, and the retouched results are used as $\{{\bm y}^{\rm new}_i\}_{i=1}^{20}$.
We enhance ${\bm x}_{\rm unseen}$ based on $\{{\bm x}^{\rm new}_i$, ${\bm y}^{\rm new}_i\}_{i=1}^{20}$ using the proposed method, the baseline method ({\it i.e.,} PieNet), and the second-best method on FiveK ({\it i.e.,} AdaInt). For further comparison, we use a general ({\it i.e.,} not personalized) enhancement method: AdaInt trained with ExpertC in FiveK, who is the most subjectively highly rated in FiveK.
The subjects are presented with two images enhanced by our method and one of the previous
methods and asked to select the better image
as shown in Figure~\ref{userstudy_vis}(d).
The voting rates are shown in Table~\ref{userstudy}. Our method is rated higher than the other methods, indicating that our results are personalized well.

In this user study,
we initialize $\{{\bm x}^{\rm new}_i\}_{i=1}^{20}$ to retouched results by ExpertC in FiveK
to reduce the retouching cost.
Because ExpertC is the most subjectively highly rated in FiveK,
the initialized results by ExpertC are visually pleasing to all users to some extent;
therefore, the subjects only need a few adjustments to the retouching parameters,
and satisfying results can be obtained in a short time.
The retouching process takes only 13 minutes on average, which means that retouching 20 images is not a difficult task for general users by using the initialization.
Note that in a practical scenario,
users can select $\{{\bm x}^{\rm new}_i\}_{i=1}^{20}$ other than the images in Figure~\ref{userstudy_vis}(a) from FiveK,
and the same initialization can be used.
If a user wants to use images not in FiveK as $\{{\bm x}^{\rm new}_i\}_{i=1}^{20}$,
we can initialize the images using general image enhancement models trained with ExpertC, which can reduce the retouching cost.

\section{Limitations}

Our method has some limitations.

{\it 1)} While PieNet~\cite{kim2020pienet} needs only preferred retouched images and does not need corresponding original images,
our method needs preferred retouched images and corresponding original images.
This is because PieNet extracts the style embeddings from only retouched images,
and our method extracts the style and content embeddings from retouched and corresponding original images.
If a user does not have original images, the user cannot use our method.
However, general image enhancement models need retouched and corresponding original images for fine-tuning,
and all high-performing methods such as AdaInt~\cite{yang2022adaint} and RCTNet~\cite{kim2021representative}
are general image enhancement models.
Based on this fact, we can conclude that retouched and corresponding original images are needed for high performance.

{\it 2)} As shown in Table~\ref{category},
if unseen images have unexpected contents, the performance of our method is degraded.
In future works, our method should be improved to generate better results for unexpected contents.

{\it 3)} In our method, all input images are resized to $512 \times 512$ px,
and our method cannot enhance higher-resolution images.
This is because we use the same architecture for the stylized enhancer as PieNet.
Recent general image enhancement models such as AdaInt~\cite{yang2022adaint} and 3DLUT~\cite{zeng2022learning}
can enhance high-resolution images in real time.
By using these architectures,
we will improve the stylized enhancer to enhance high-resolution images.

\section{Conclusions}
We addressed PIE in this study.
To achieve content-aware PIE, we proposed masked style modeling, where the Transformer encoder receives style embeddings and content embeddings and predicts the style embedding for an unseen input.
To allow this model to consider the contents of images, we proposed the novel training scheme,
where we use multiple real users’ preferred images on Flickr.
The proposed method successfully achieved content-aware PIE. Our evaluation results demonstrated that the proposed method outperformed previous methods in both the quantitative evaluation and the user study.

\vspace{2mm}
\noindent
{\bf Acknowledgements}~
This research is approved by the Institutional Review Board (IRB) of
Graduate School of Information Science and Technology,
The University of Tokyo (UT-IST-RE-201207-1a).



   \begin{IEEEbiography}[{\includegraphics[width=1in,height=1.25in,clip,keepaspectratio]{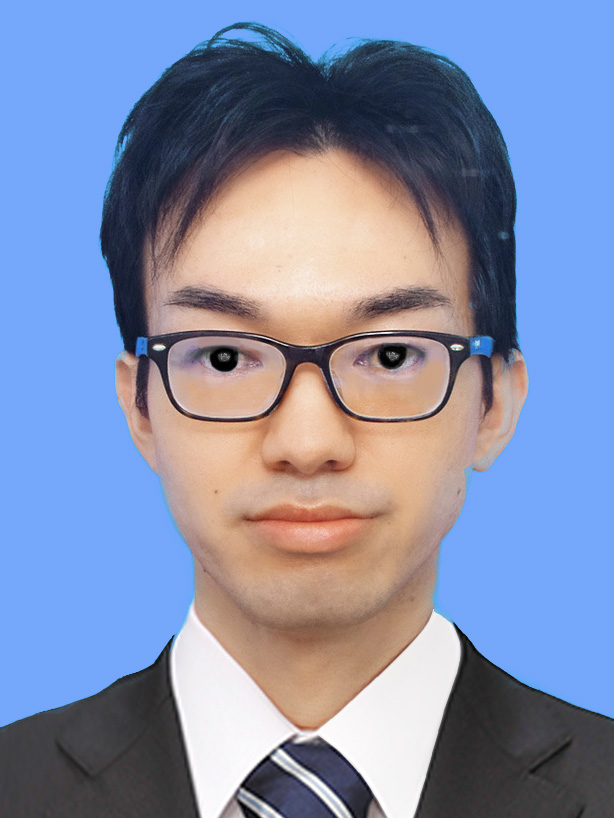}}]{Satoshi Kosugi}
     received the B.S., M.S., and Ph.D. degrees in information and communication
engineering from The University of Tokyo, Japan,
in 2018, 2020, and 2023, respectively.
He is currently an Assistant Professor at
the Laboratory for Future Interdisciplinary Research of Science and Technology,
Institute of Innovative Research, Tokyo Institute of Technology.
His research interests
include computer vision, with particular interest in
image enhancement.
   \end{IEEEbiography}

   \begin{IEEEbiography}[{\includegraphics[width=1in,height=1.25in,clip,keepaspectratio]{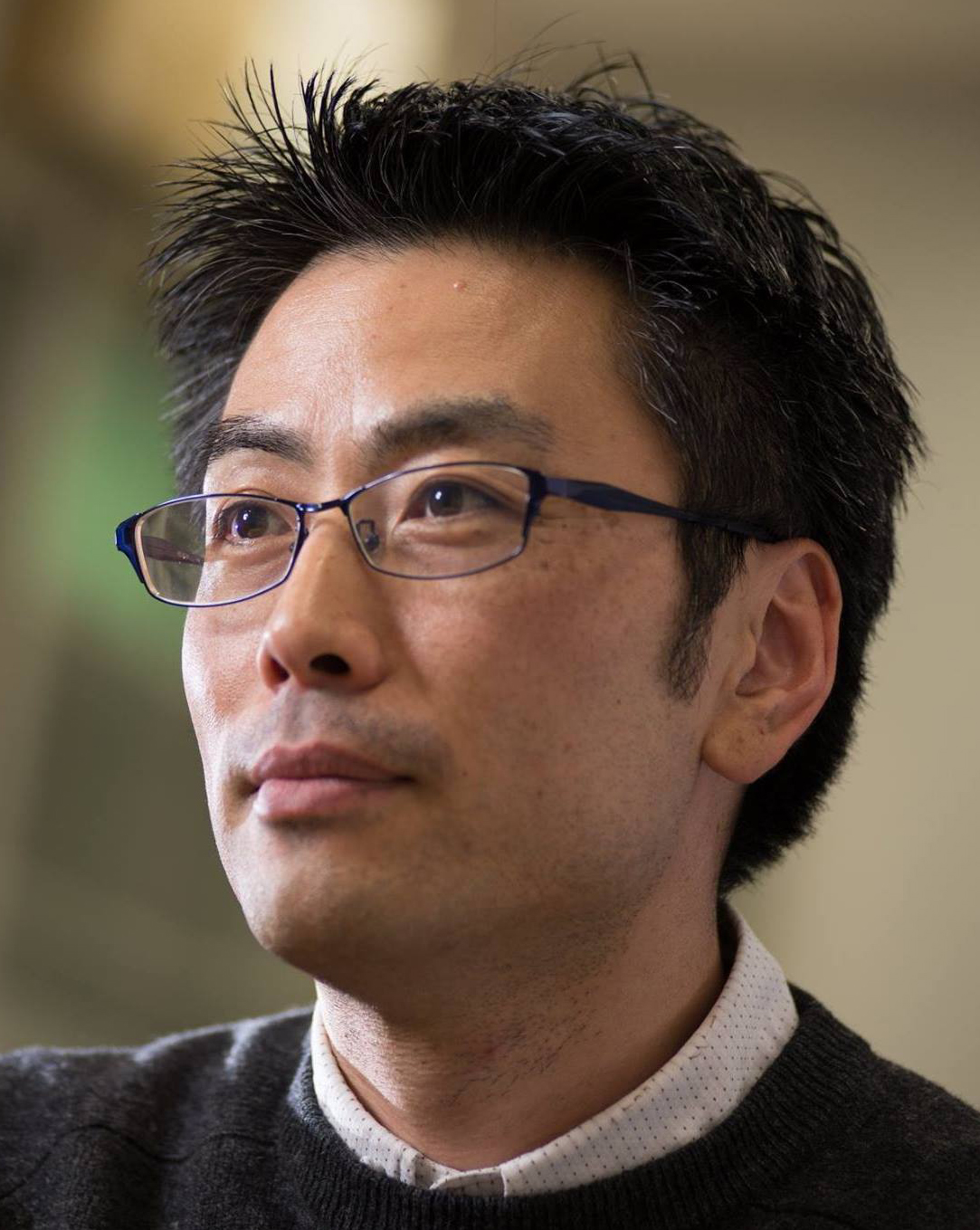}}]{Toshihiko Yamasaki}
   (Member, IEEE) received the
B.S., M.S., and Ph.D. degrees from The University
of Tokyo. He was a JSPS Fellow for Research
Abroad and a Visiting Scientist at Cornell University,
Ithaca, NY, USA, from February 2011 to February
2013. He is currently a Professor at
the Department of Information and Communication Engineering, Graduate School of Information
Science and Technology, The University of Tokyo.
His current research interests include attractiveness
computing based on multimedia big data analysis
and fundamental problems in multimedia, computer vision, and pattern recognition.
He is a member of ACM, AAAI, IEICE, IPSJ, JSAI, and ITE.
\end{IEEEbiography}

\end{document}